\documentclass[runningheads]{llncs}
\usepackage{listings}
\usepackage{xcolor}
\usepackage{tcolorbox}
\usepackage{graphicx}
\usepackage{cite}
\usepackage{verbatim}
\usepackage{multirow}
\usepackage{amsmath}
\usepackage{textcomp}
\usepackage{caption,subcaption}
\usepackage{verbatim}
\usepackage{changepage}
\usepackage{makecell}
\usepackage{threeparttable}
\usepackage[euler]{textgreek}
\usepackage{amssymb}
\usepackage[ruled,vlined]{algorithm2e}
\usepackage{graphicx}
\begin{document}
\title{Building a Competitive Associative Classifier}
%
%
\author{Nitakshi Sood \and
Osmar Zaiane\orcidID{0000-0002-0060-5988}}
%
\authorrunning{Sood et al.}
%
\institute{Alberta Machine Intelligence Institute\\University of Alberta, Edmonton AB T6G2R3, Canada 
\email{\{nitakshi,zaiane\}@ualberta.ca}}
\maketitle              
\begin{abstract}
With the huge success of deep learning, other machine learning paradigms have had to take back seat. Yet other models, particularly rule-based, are more readable and explainable and can even be competitive when labelled data is not abundant. However, most of the existing rule-based classifiers suffer from 
the production of 
a large number of classification rules, affecting the model readability. This hampers the classification accuracy as noisy rules might not add any useful information for classification and also lead to longer classification time. In this study, we propose SigD2 which uses a novel, two-stage pruning strategy which prunes most of the noisy, redundant and uninteresting rules and makes the classification model more accurate and readable. 
To make SigDirect more competitive with the most prevalent but uninterpretable machine learning-based classifiers like neural networks and support vector machines, we propose bagging and boosting on the ensemble of the SigDirect classifier. The results of the proposed algorithms are quite promising and we are able to obtain a minimal set of statistically significant rules for classification without jeopardizing the classification accuracy. We use 15 UCI datasets and compare our approach with eight existing systems. The SigD2 and boosted SigDirect (ACboost) ensemble model outperform various state-of-the-art classifiers not only in terms of classification accuracy but also in terms of the number of rules.

\end{abstract}

\section{Introduction}
Classification is defined as a process of predicting the class label of new data points, given a set of labeled data points in the training set. The association rule mining is a rule-based approach that helps in identifying patterns in the data in the form of rules, by finding the relationships between the items in the dataset. The association rules are in the form X \textrightarrow Y, where X is the antecedent and Y is the consequent \cite{agrawal1994fast}. Associative classifiers combine the concept of association rule mining and classification to build a classification model. In an associative classifier, we choose the consequent of the rule to be the class label and the antecedent set is a set of attribute-value pairs for the associated class label. In the literature, various associative classifiers have been proposed till now namely, CBA \cite{Liu:1998:ICA:3000292.3000305}, CMAR \cite{cmarpaper}, CPAR \cite{doi:10.1137/1.9781611972733.40},  ARC\cite{1183881} etc. Although these classifiers are easily understandable, flexible and do not assume independence among the attributes, they require prior knowledge for choosing appropriate parameter values (support and confidence). 
Furthermore, the rules generated 
may include noisy and meaningless rules, which might hinder the classification. A rule is said to be noisy if it does not add any new information for prediction and instead misleads the classification model. In other terms, a noisy rule would participate more often in misclassifications than in correct classifications. 

The authors in \cite{li2017exploiting} proposed SigDirect, an associative classifier which mines statistically significant rules without the need for the support and confidence values. 
However, in this paper, we propose SigD2 where we introduce a more effective two stage pruning strategy to obtain a more accurate classification model. The proposed method reduces the number of rules to be used for classification without compromising on the prediction performance. In fact, the performance is improved. Most of the prevalent supervised classification techniques like Artificial Neural Networks (ANN), Support Vector Machines (SVM) etc, although provide very high classification accuracy, they act as a black box. The models produced by such classifiers are not straight forwardly explainable. However, the proposed associative classifier makes the model more explainable by 
producing only a minimal set of classification association rules (CARs). The proposed technique finds its immense usage in various health-care related applications, where the explanation of proposed models along with the classification accuracy are highly significant~\cite{JingZhang2013}. In health-care, incorrect predictions may have catastrophic effect, so doctors find it hard to trust AI unless they can validate the obtained results.

Furthermore, we also propose ACboost, which uses an ensemble of classification models obtained from the weak version of SigDirect, for boosting. Our goal is to strengthen the classifier using less number of rules for prediction. Since, SigDirect is a strong learner and produces already a lesser number of rules for prediction, we form a weak version of SigDirect called wSigDirect, by further reducing the number of rules to be used for classification as explained later in Section \ref{sec:method}. Moreover, in the proposed approach we use Adaboost \cite{freund1996experiments} based boosting strategy over the ensemble of wSigDirect. The wSigDirect's classification model is learnt by running it multiple times on a re-weighted data, thereafter performs voting over the learned classifiers. 
We also propose ACbag which is defined as bagging on an ensemble of wSigDirect classifiers. Motivated by the approach proposed by Breiman in \cite{breiman1996bagging}, we use an ensemble model of wSigDirect classifiers trained in parallel over different training datasets, and perform a majority voting over the ensemble for prediction. With the use of this strategy of combining weak learners, the goal is to decrease the variance in the prediction and improve the classification performance henceforth.  

It was found that for most of the datasets ACboost performs better than SigD2, ACbag, SVM, or ANN; ANN which performs similarly to deep neural network on these reasonably sized datasets. The main aim of this study is to make associative classifiers more competitive and to highlight their significance as opposed to the other machine learning based classifiers like neural networks which do not produce explainable predictions. Deep Learning has garnered all the attention lately, but the inability to produce transparent explanations for the decisions motivates us towards the domain of explainable artificial intelligence using rule-based models which have fallen out of favour of late.

Our contribution in this study is as follows:
\vspace{-1mm}
\begin{itemize}
\setlength{\itemsep}{1pt}
\setlength{\parsep}{1pt}
\setlength{\parskip}{1pt}
    \item We propose SigD2, an associative classifier, which uses an effective two stage pruning strategy for pruning the rules to be used for classification. Using the proposed approach, the number of rules used for classification are reduced notably, without compromising on the classification performance.
    \item We propose ACbag, an ensemble based classifier founded on wSigDirect.
    \item We also propose ACboost, which is boosting the wSigDirect classifier, to improve the classification accuracy with an explainable base model. Therefore, making SigDirect more competitive for classification tasks.
\end{itemize}

The rest of the paper is organized as follows: Section~\ref{sec:related} gives a literature review about some previously proposed associative classifiers, Section~\ref{sec:method} explains the methodologies we have adapted in SigD2, ACbag and ACboost, Section~\ref{sec:eval} shows the evaluation results of our proposed classifier on UCI datasets and lastly, Section~\ref{sec:conclusion} gives the conclusion of the work and directions about future investigations.

\section{Related Work}\label{sec:related}
In this section, we briefly describe some related work on associative classification. Stemming from association rule mining, associative classifiers have been extensively studied in the last two decades. 
Liu et al. first proposed the classification based on association (CBA) technique in \cite{agrawal1994fast} and showed that the association rule mining techniques could be applicable to classification tasks. CBA uses the Apriori algorithm to generate Classification Association Rules (CARs) and database coverage for pruning the noisy rules. It uses the highest ranked matching rules as the heuristic for classification. Inspired by the idea of CBA, many authors came up with more efficient versions of associative classifiers. CPAR proposed by Yin and Han uses a dynamic programming based greedy strategy that generates association rules from the training dataset\cite{doi:10.1137/1.9781611972733.40}. It prevents repeated calculation in rule generation and also selects best $k$ rules in prediction. 
CMAR proposed by Li et al. uses an FP growth algorithm \cite{cmarpaper} to produce a set of CARs which are stored in a tree-based data structure called CR-tree. They also use database coverage for rule pruning and finally make a prediction based on the multiple matching rules with a weighted chi-square measure.
The ARC model \cite{1183881} takes all the rules which lie within the confidence range, then calculates the average confidence for the rules grouped by the class labels. The class label of the group with the highest confidence average is finally selected for prediction. The CCCS algorithm in \cite{Arunasalam:2006:CTA:1150402.1150461} is proposed for imbalanced dataset classification problems where using support/confidence framework would not be sufficient. The classification using complement class support algorithm mines the positively correlated CARs by using a novel measure called complement class support(CCS) conjointly with top-down row enumeration algorithm. However, CCCS does not assure the statistical significance of the mined association rules, which defeats the purpose we would like to achieve. Moreover, they only compare their approach with the original CBA. 

Antonie et al. proposed a two stage classification model called 2SARC in \cite{antonie2006learning} which automatically learns to select appropriate rules for classification. In the first stage, association rule mining is used to determine the classification rules. These rules are further used to determine meaningful features to be used for predicting which rules are to be selected for induction. In the second stage, these multiple features are given as input to another learning algorithm that is, a neural network, in order to obtain a more accurate classification model with high prediction accuracy. However, the model produced by the proposed approach does not seem to give explainable results.
Li and Zaiane presented a novel associative classifier in \cite{li2015associative} which is built upon both positive and negative association classification rules.
They improvised the kingfisher algorithm for rule generation, and also proposed a novel pruning strategy for both positive and negative rules simultaneously. The authors state that a generated rule can be pruned if it is found to incorrectly classify at least one training instance. Finally in the classification stage, they concluded that summing up the confidence values of all matching rules and accordingly making the class label prediction proves to be the best classification method.

The associative classifiers have the ability to provide a readable classification model. The study done in \cite{zaiane2005pruning} focuses on the significance of obtaining a minimal set of CAR's without jeopardising the performance of the classifier. They propose a pruning strategy to reduce the number of rules in order to build an effective classification model without seriously compromising on the classification accuracy. The authors also propose heuristics to select rules which obtain high accuracy on the plot of correct/incorrect classification for each rule on the training set for effective rule pruning combined with the database coverage technique based on the given dataset. Tuning values for support and confidence parameters is an arduous task as it varies with the change in dataset. Li and Za\"{\i}iane in \cite{li2017exploiting} overcome this limitation by proposing SigDirect that tunes only one parameter that is the p-value, which computes the statistical significance of rules using Fisher's exact test. The authors proposed an instance centric rule pruning strategy for pruning the non statistically significant rules. Although SigDirect has proved to be quite competitive in terms of prediction, there are still noisy rules that can compromise the accuracy.

Furthermore, ensemble models are widely used for enhancing the accuracy of the classification models using a combination of weak learners. The SAMME algorithm proposed by Hastie et al. in \cite{hastie2009multi} is a multi-class extension of the binary Adaboost algorithm \cite{freund1996experiments}. 

\section{Methodology}\label{sec:method}
In this section, we introduce the details about the proposed effective pruning technique as used in SigD2. Further, we extend our work to perform bagging and boosting over the ensemble of wSigDirect associative classifier.  
\subsection{SigD2}
The aim of an associative classifier is to find knowledge from data in the form of association rules associating conjunctions of attribute-value pairs with class labels, and then use these rules for prediction. SigD2 processes the learning of rules in rule generation and rule pruning phases. It further uses these rules for prediction in the classification phase.
\subsubsection{Rule Generation phase:}
In this phase, we use the approach proposed by Li and Za\"{\i}ane for SigDirect in \cite{li2017exploiting}. SigD2 generates statistically significant CARs, such that the p-value from Fisher's exact test \cite{hamalainen2010efficient} of the rule in the form X \textrightarrow c\textsubscript{k} is small.  
Initially all the impossible antecedent itemsets are removed using the corollary defined in \cite{li2017exploiting}. The remaining items are sorted and arranged in the ascending order of their support values. The enumeration tree is built over the remaining items. For the first level, all the CARs with one antecedent values are listed and checked if they are potentially statistically significant (PSS) \cite{li2017exploiting}. All the non PSS CARs are pruned from the tree while, for the CARs which are PSS, exact p-value is calculated to find out if it is statistically significant. Using 1-itemset PSS CARs from the first level, 2-itemset PSS CARs are generated and this process is repeated until a certain level is reached where no PSS CARs can be generated. Furthermore, it is also checked if the p-value of a CAR is smaller than a significance level of 0.05 and the CAR is non redundant and minimal \cite{li2017exploiting}.
The CAR in the form X\textrightarrow c\textsubscript{k} is said to be non redundant if there does not exist any CARs in the form of y \textrightarrow c\textsubscript{k}, such that p(y\textrightarrow c\textsubscript{k}) \textless p(X\textrightarrow c\textsubscript{k}) and y is proper subset of X, where p is the p-value of the rule calculated from Fisher's exact test, X is the set of items in the database and C\textsubscript{k} is the class label \cite{li2017exploiting}. While a CAR is termed as minimal if X\textrightarrow c\textsubscript{k} is non redundant and there does not exist any CARs in the form Z\textrightarrow c\textsubscript{k} such that X is a proper subset of Z and p(Z\textrightarrow c\textsubscript{k}) \textless p(X\textrightarrow c\textsubscript{k}).
So, if a CAR is found to be minimal then it is impossible for all its children in the subtree to get a lower p-value \cite{li2017exploiting}. Therefore, the tree is not enumerated further. 

\begin{algorithm}[htb]
\SetAlgoLined
{\textbf{Input:} \textbf{T}: Pruning transaction database, \textbf{R}: Initial rule list from rule generation phase, \textbf{ R\textsubscript{mid}}: Rule list being formed after pruning the insignificant rules from R, \textbf{conf\_threshold}: Confidence threshold value.}

\KwResult{\textbf{ R\textsubscript{new}}: Classification association rules to be used for prediction }
\While{rules exist in R}{
Sort the rules in R in descending order of their confidence values \\
Select the rule r\textsubscript{i} with highest confidence from R and add to the R\textsubscript{mid} \\ 
\uIf{conf(r\textsubscript{i}) $\textless$ conf\_threshold} {break}
Find all applicable instances in T that match the antecedent of rule r\textsubscript{i}  \\
\uIf{r\textsubscript{i} correctly classifies a pruning instance in T}{
Mark r\textsubscript{i} as a candidate rule in the classifier \\
Remove all instances in T covered by r\textsubscript{i} } 
Update the confidence values, based on the remaining transactions\\
Remove the rule r\textsubscript{i} from the R
}
\For{each instance t in the original transaction database T}{
Scan the CARs from R\textsubscript{mid} to find the matching CAR r\textsubscript{i}, with highest confidence value \\
\eIf{r\textsubscript{i} $\not\in$ R\textsubscript{new}} { R\textsubscript{new}.add(r\textsubscript{i})\\
		r\textsubscript{i}.count=1
}{r\textsubscript{i}.count+=1}
}
    \caption{Algorithm for Two-Stage Pruning Strategy used in SigD2}
\end{algorithm}

\subsubsection{Rule Pruning Phase:}
The rule generation phase may produce many CARs which are noisy and would not only slow down the process of classification but also lead to incorrect classification. Originally, SigDirect only performs instance based rule pruning on generated rules. It was observed that, although the previous strategy produces globally best CARs, the rules were still noisy and could be further reduced. So the question is, how can we prune more rules without  actually jeopardising the accuracy of the associative classifier?


We propose a two stage strategy for pruning, wherein we randomly divide the training set into train set and prune set in the ratio of 2:1. The rules are generated in the rule generation phase using the train set. However, for pruning, only the prune set is used. We sort the CARs in the descending order according to confidence values. The proposed pruning process, consists of matching the CAR with highest confidence and scanning over all the transactions in the pruning dataset to see if they match. 
If the rule applies correctly on the transactions, it is marked and is selected to be used for classification and subsequently the matching transactions are removed from the pruning set. We re-calculate the confidence values of the remaining rules, each time using the remaining transactions in the pruning set and arrange them in the descending order. This process is repeated until either the rules or transactions have been covered or until the confidence threshold is reached. It is assumed that for a rule, if the confidence value in each iteration is less than the threshold, then that rule can be pruned as it is not able to cover at least few instances in the prune set.



After this step, we obtain the rules which might be useful for classification. However, we still need to find the globally best CARs. So, further we apply the instance based pruning step as proposed in SigDirect \cite{li2017exploiting}. For every instance in the pruning transaction database, the complete set of CARs generated from the previous step are scanned. The aim here is to find the matching CARs with the highest confidence value, such that, the class label of the rule and the transaction matches and the antecedent of the rule is the subset of 
the transaction. Furthermore, the count of how many times the rule has been selected in the pruning instances is maintained. This is later used in order to perform weighted classification using the number of times the CAR was selected in the pruning phase. 
Using the proposed approach only high quality rules with high confidence values are kept. 

\begin{figure}[!h]
\begin{center}
\includegraphics[width=0.95\textwidth, height = 4.85cm]{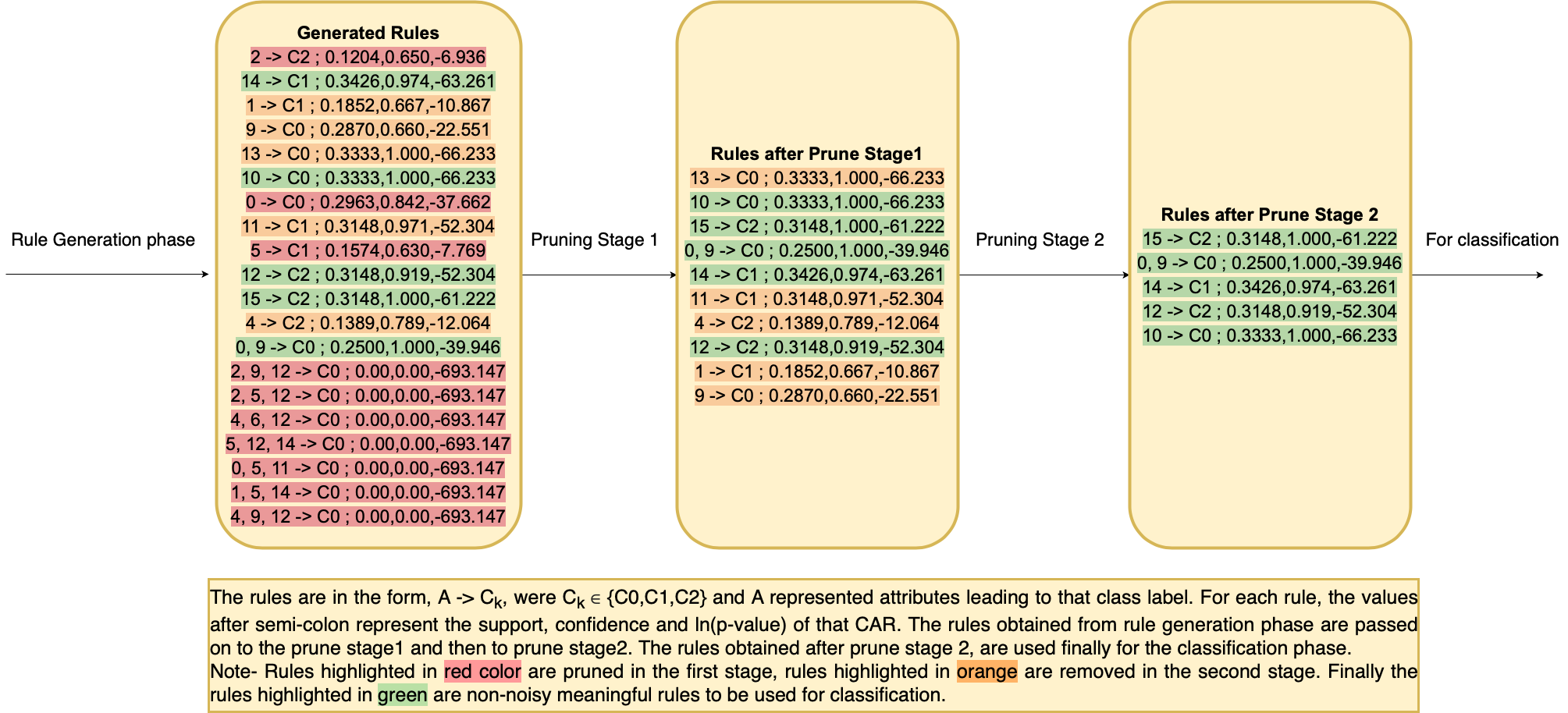} 
\caption{An example illustrating the two stage pruning process}
\label{fig:2stageprune}
\end{center}
\end{figure}

High quality rules are the non noisy rules which do not make mistakes on the pruning set. 
Figure \ref{fig:2stageprune} shows an example of two stage pruning process on the Iris dataset. The first block in the figure shows all the rules obtained from the rule generation phase. All the rules highlighted in red, are the ones that are pruned in the first stage of pruning itself for being noisy and meaningless. Further the rules highlighted in orange, are the ones that are pruned in the second stage of pruning, for not being globally optimal. Finally, after pruning out all the noisy CARs, the remaining rules are the high quality rules, that are highlighted in the green color and are further used in the classification phase. This pruning  strategy also avoids over-fitting on the data. 

\subsubsection{Classification Phase:}
After the pruning phase, the minimal set of statistically significant rules are obtained. Further we make predictions on the new instances from the test set. For a given new instance, the classification process would search the subset of the CARs that match the new instance in order to predict its class label. All the matching CARs are then divided into groups based on their class labels. Then we use the three heuristics proposed in \cite{li2017exploiting}, which state that each class's group should be ordered on the basis of sum of ln(p-value), sum of confidence value and the sum of ln(p-value).confidence. The class label for the group sorted on the values of sum of ln(p-value) is obtained from the class of the matching CAR with the lowest ln(p-value). Moreover, the class label for the group sorted on values of sum of ln(p-value).confidence is obtained from the class of the matching CAR with lowest ln(p-value).confidence value. Finally, the class label for the group sorted on the values of sum of confidence value is determined from the class of the CAR with the highest confidence value. Furthermore, we can also use two stage classification as proposed by Antonie et al. in \cite{antonie2006learning}, to learn with a neural network in a second phase to predict the the classification rules to use.

\subsection{Bagging and Boosting on wSigDirect} 
In this section, we perform bagging and boosting on the weak version of SigDirect, we call wSigDirect. While SigDirect is already a strong learner, we chose it over CBA as it gives a smaller number of rules. But we need to make it weaker to be used for ACbag and ACboost. We do this by further reducing the number of rules to be used for classification.
The strategy for rule generation and rule pruning stays similar to that of the original SigDirect. However, for all the association rules obtained from the pruning phase for classification, we divide these rules as per the class label. Further, we chose the top $\eta$ rules on the basis of highest confidence values from each class label group. The classification model thus obtained is called weak as it does not involve all the significant rules. We perform bagging and boosting on the ensemble model of wSigDirect over different trained datasets for prediction. 
\subsubsection{Bagging:}
ACbag is motivated by the approach proposed in \cite{breiman1996bagging}. The weak classifiers are learnt in parallel by picking instances randomly with replacement from the training data. Each wSigDirect model is learnt independent of each other. In bootstrap sampling, every observation has equal probability of appearing in the training dataset. Finally, we perform a majority voting over the results of the weak learners and predict the class label for each testing sample. This approach helps in avoiding the problem of overfitting. Since, the base models are explainable, the ACbag can explain the responses of each learner, and the explanation of the ensemble would be the set of rules that were voted on by the ensemble. Furthermore, it was observed that the results obtained after performing bagging on wSigDirect are very comparable or slightly better than those achieved by bagging on the original SigDirect. Therefore, later in Section 4, we report the results of bagging on wSigDirect.
\subsubsection{Boosting:}
Boosting is a process of improving the performance of a weak learning algorithm. It is done under the assumption that, the performance of the weak learner is at least slightly better than random guessing on different observations. 
In this phase, we propose ACboost which iteratively calls wSigDirect. This weak learner is converted to a strong learner either by weighted average of the predictions from weak learners or by considering prediction with majority voting. Given a training set, with features and class labels, we initialize the weights of our samples as one divided by the number of training instances.
For the number of weak learners to be used sequentially, we train the first base learner using wSigDirect and obtain the misclassification error of the model. Further, the weight of the classifier is calculated based on its performance on the training data. Finally, the weight of each sample is updated, such that samples that were correctly classified are given less weights whereas the samples which were incorrectly predicted are given more weights. This would force the learner to pay more attention towards the incorrect predictions done by the previous learner.
The iteration is continued till the maximum number of estimators (pre-set number of weak learners) are reached or a low training error is achieved. Finally, the prediction is done by using the weights of each classifier calculated previously to perform weighted prediction. This sequential learning of models helps in reducing the training error. We have used the methodology proposed for multi-class classification in SAMME algorithm \cite{hastie2009multi}, an extension of adaboost, which adds up a log term to the weight of the classifier making the boosting algorithm applicable for both two-class and multi-class classification tasks. Furthermore, since the rules produced by the base classifier are explainable therefore, there is a possibility of interpretation of results.


\section{Experimental Results}\label{sec:eval}
We evaluate our SigD2 associative classifier on 15 UCI datasets \cite{Dua:2019}. We discretize the datasets as proposed in \cite{CF_2004}, so the classification accuracy  might be marginally different from the previously reported results. We report the results after performing the average over 10 fold cross validation on each dataset. We use 90\% of the total data as the train set and further divide the train set into train set and prune set in the ratio of 2:1.
\begin{table*}[htb] 
    \centering
    \caption{Comparison of classification accuracy of SigD2 with other rule-based classifiers}   
    \label{tab:accuracy_phase1}
\begin{scriptsize}
\begin{tabular}{|p{1.5cm}||p{0.65cm}|p{0.75cm}||p{0.75cm}|p{1cm}|p{1cm}|p{1 cm}|p{1.25 cm}|p{1.5cm}|p{1.5cm}|}
\hline 
\textbf{Datasets}&\textbf{\#cls}&\textbf{\#rec}&\textbf{C4.5}&\textbf{CBA}&\textbf{CMAR}&\textbf{CPAR}&\textbf{RIPPER}&\textbf{SigDirect}&\textbf{SigD2}\\

\hline
Adult&2&48842&78.8&\textbf{84.2}&81.3&77.3&\textbf{84.1}&\textbf{84.1}&83.59\\
\hline
Anneal&6&898&76.7&94.5&90.7&95.1&\textbf{98.32}&96.99&\textbf{97.21}\\
\hline
Breast&2&699&91.5&\textbf{94.1}&89.9&93&\textbf{95.42}&91.7&92.7\\
\hline
Flare&9&1389&82.1&84.2&\textbf{84.3}&63.9&72.13&\textbf{84.23}&\textbf{84.3}\\
\hline
Glass&7&214&65.9&68.4&\textbf{71.1}&64.9&68.69&\textbf{70.56}&69.17\\
\hline
Heart&5&303&\textbf{61.5}&57.8&56.2&53.8&53.97&58.49&\textbf{59.81}\\
\hline
Hepatitis&2&155&84.1&42.2&79.6&75.5&78.06&\textbf{85.83}&\textbf{86}\\
\hline
Horse&2&368&70.9&78.8&82.3&81.2&\textbf{84.23}&81.23&\textbf{85.03}\\
\hline
Iris&3&150&91.3&93.3&94&94.7&\textbf{95.33}&94&\textbf{96}\\
\hline
Led7&10&3200&\textbf{73.9}&73.1&73.2&71.3&69.15&73.78&\textbf{73.81}\\
\hline
Mushroom&2&8124&92.5&46.5&\textbf{100}&\textbf{98.5}&\textbf{100}&\textbf{100}&\textbf{100}\\
\hline
PageBlocks&5&5473&92&90.9&90.1&\textbf{92.5}&\textbf{96.83}&91.21&92.18\\
\hline
Pima&2&768&70.5&74.6&74.4&74&66.36&\textbf{75.25}&\textbf{74.86}\\
\hline
Wine&3&178&71.7&49.6&92.7&88.2&91.57&\textbf{92.71}&\textbf{93.2}\\
\hline
Zoo&7&101&91&40.7&\textbf{93}&\textbf{94.1}&87.12&91&89.18\\
\hline \hline
Average&&&79.62&71.52&83.52&81.2&82.75&\textbf{84.73}&\textbf{85.13}\\
\hline

\end{tabular}
\begin{tablenotes}
      \small
      \centering
      \item Note- \#cls indicates number of class labels and \#rec indicate the number of records in dataset.
\end{tablenotes}
\end{scriptsize}
\end{table*}

\subsection{Classification Accuracy}
We compare the performance of the proposed classifiers on 15 UCI datasets, with other rule-based classifiers like CBA, CMAR, CPAR, RIPPER, C4.5 and the original SigDirect, in terms of classification accuracy and number of classification rules in the final model. Further, we also compare ACboost with ANN and SVM in Table \ref{tab:accuracy_phase2}. We use the default parameters as stated by the authors in original respective papers as well as stated in \cite{li2017exploiting}. In CBA and CMAR the parameters are tuned such that the minimum confidence values is set to be 50\% , minimum value of support is set as 1\%, the maximum number of CARs are limited to 80,000 and the size of number of  antecedent items are limited to 6. In CPAR, the minimum gain threshold is set to 0.7, decay factor to 2/3 while the threshold for the total weight is set to be 0.05. For RIPPER\cite{cohen1995fast}, we use default JRip from WEKA \cite{holmes1994weka}. The default parameters as stated in the original papers are used for C4.5 \cite{Salzberg1994}, SVM \cite{cortes1995support} and SigDirect \cite{li2017exploiting}.

For SigD2, we have performed a sensitivity analysis on the confidence threshold and it was found that threshold value lower than 30\% or higher than 50\%, does not lead to best results for all the considered datasets. Hence, we chose to vary the confidence threshold in the range of 30-50\% depending on the dataset.
\begin{table*}[t]
    \centering
    \caption{Comparison of classification accuracy of ACboost with ACbag, SigD2, SigDirect, ANN and SVM}   
    \label{tab:accuracy_phase2}
\begin{scriptsize}
\begin{tabular}{|p{1.5cm}|p{1cm}|p{1cm}|p{1cm}|p{1.5cm}|p{1.5cm}|p{1.5cm}|p{1.5cm}|}
\hline 
\textbf{Datasets}&\textbf{SVM}&\textbf{ANN}&\textbf{DNN}&\textbf{SigDirect}&\textbf{SigD2}&\textbf{ACbag}&\textbf{ACboost}\\
\hline 
Adult&75.8&75.66&\textbf{85.35}&84.1&83.59&84.74&\textbf{85.23}\\
\hline
Anneal&85&93.964&\textbf{97.6}&96.99&97.21&\textbf{97.43}&97.31 \\
\hline
Breast&95.7&\textbf{96.83}&\textbf{96.48}&91.7&92.7&93.86&92.62\\
\hline
Flare&73.8&\textbf{84.61}&70.3&84.23&84.3&84.31&\textbf{85.35}\\
\hline
Glass&68.6&70.148&66.9&70.56&69.17&\textbf{72.01}&\textbf{76.96}\\
\hline
Heart&55.4&56.72&55.6&58.49&59.81&\textbf{61.33}&\textbf{63.74}\\
\hline
Hepatitis&79.3&82.89&83.07&85.83&\textbf{86}&85.18&\textbf{90.89}\\
\hline
Horse&72.5&81.321&80.9&81.23&85.03&\textbf{85.3}&\textbf{85.7}\\
\hline
Iris&94.6&\textbf{98.09}&95.8&94&96&94.66&\textbf{97.33}\\
\hline
Led7&73.6&69.64&68.63&73.78&73.81&\textbf{74.84}&\textbf{75.21}\\
\hline
Mushroom&\textbf{99.8}&\textbf{100}&\textbf{100}&\textbf{100}&\textbf{100}&\textbf{100}&\textbf{100}\\
\hline
PageBlocks&91.2&\textbf{95.42}&95.08&91.21&\textbf{92.18}&91.24&92.13\\
\hline
Pima&74&\textbf{75.95}&75.15&75.25&74.86&75.53&\textbf{75.55}\\
\hline
Wine&94.9&91.662&\textbf{97.62}&92.71&93.2&94.04&\textbf{98.85}\\
\hline
Zoo&92.2&93.192&89.94&91&89.18&\textbf{94.28}&\textbf{98.9}\\
\hline \hline 
Average&81.76&84.406&83.89&84.738&85.136&\textbf{85.91}&\textbf{87.71}\\
\hline

\end{tabular}

\end{scriptsize}
\end{table*}
For ANN, we use a shallow network with one hidden layer. The number of nodes in the hidden layer are set as the average of number of input and output nodes. The architecture may vary slightly with dataset, but
we use ReLU (Rectified Linear Units) or sigmoid functions for activation and around 200 training epochs with a learning rate of 0.1.
For ACboost and ACbag, the value of $\eta$ is tuned in the range of 5-15 for every dataset. The number of estimators are varied in the range of 15-100 for each fold in every dataset and we report the best results.  The value for parameters $\eta$ and the number of estimators have been concluded after performing a sensitivity analysis on each of them.

\begin{table*}[htb]
    \centering
    \caption{SigD2 compared with other algorithms on the basis of number of rules}   
    \label{tab:rules}
\begin{scriptsize}
\begin{tabular}{|p{1.5cm}|p{1cm}|p{1cm}|p{1cm}|p{1cm}|p{1.5cm}|p{1.5cm}|p{2.65cm}|}
\hline 
\textbf{Datasets}&\textbf{C4.5}&\textbf{CBA}&\textbf{CMAR}&\textbf{CPAR}&\textbf{SigDirect}&\textbf{SigD2}&\textbf{Difference with Average \# of rules}\\
\hline  
Adult&1176.5&691.8&2982.5&84.6&91.2&53.62&951.7 (94.67\%)\\
\hline
Anneal&17&27.3&208.4&25.2&41.7&29.2&34.72 (54.31\%)\\
\hline
Breast&8.8&13.5&69.4&6&10.9&7&14.72 (67.65\%)\\
\hline
Flare&54.4&115.1&347.1&48.1&75.8&25.7&102.4 (79.93\%)\\
\hline
Glass&14.8&63.7&274.5&34.8&55.6&23.1&65.58 (73.9\%)\\
\hline
Heart&23.9&78.4&464.2&44&80.2&27.7&110.44 (77.3\%)\\
\hline
Hepatitis&8.1&2.3&165.7&14.3&33.3&16&28.74 (64.23\%)\\
\hline
Horse&25.6&116.4&499.9&19&90.4&41.5&108.76 (72.38\%)\\
\hline
Iris&8.4&12.3&63.4&7.4&6.2&4.8&14.74 (75.43\%)\\
\hline
Led7&63.2&71.2&206.3&31.7&104.3&54.4&40.94 (42.94\%)\\
\hline
Mushroom&121.2&2&102.6&11.1&106.4&48.9&19.76 (28.77\%)\\
\hline
PageBlocks&16.3&7.6&80.6&29.9&31.1&13.2&19.9 (60.12\%)\\
\hline
Pima&24.4&43.2&203.3&21.7&36.6&11.3&54.54 (82.83\%)\\
\hline
Wine&12.8&4.7&122.7&15.2&29.3&16.3&20.64 (55.87\%)\\
\hline
Zoo&5.3&2&35&16.9&16.2&9&6.08 (40.31\%)\\
\hline 

\end{tabular}
\end{scriptsize}
\end{table*}
Table \ref{tab:accuracy_phase1} shows that SigD2 performs quite well as compared to other rule-based and associative classifiers. The average performance over 15 datasets of SigD2 is better than all the other rule-based classifiers. Although, the difference between SigDirect and SigD2 on the basis of classification accuracy is marginal, when we compare the number of rules, we show that SigD2 outperforms SigDirect. In order to have a fair comparison, among different algorithms on various datasets, we analyse how many times did an algorithm win and how many times it was a runner up as shown in Table \ref{tab:bestrunnerup}. The proposed pruning strategy is found to give quite promising results as compared to the other rule-based and associative classifiers. SigD2 outperforms RIPPER on 10 out of 15 datasets. Furthermore, Table \ref{tab:accuracy_phase2} shows that ACboost outperforms all the classifiers including SigDirect, SigD2, ANN and SVM. We have also tried to compare our approach with deep neural network (DNN) with 5 hidden layers. ACboost was found to perform better than DNN in 10 with 1 tie out of 15 datasets. However, since most of the considered datasets are not big enough to be used for DNN, the results might not be conclusive.
\subsection{Number of Rules}
The main advantage of the associative classifiers over the other machine learning supervised classifiers is its ability to build a model which is human readable. 
Noisy, redundant and uninteresting rules lead to longer classification time, reduce the performance of the classifier and also make it tedious for humans to analyse the model. Ideally, we want to achieve maximum accuracy with a minimum possible set of rules. Table \ref{tab:rules} shows the comparison among different classifiers on the basis of number of rules generated. The two stage pruning technique is found to give a minimum number of rules without compromising the classification performance. Table \ref{tab:pvalue} clearly shows that out of 15 datasets, on average SigD2 outperforms most of the other rule-based and associative classifiers for at least 10 datasets with some ties in few cases as well. 

SigD2 gets a smaller number of rules on 9 datasets when compared with CBA. Although, CBA is found to have less rules for some datasets, it is unable to provide a high accuracy in such cases. Our proposed strategy outperforms CMAR on all datasets, the original SigDirect on all but one dataset and CPAR, C4.5 on 8 datasets. The number of rules is found to be appropriate enough to provide information about the classification model without compromising on the performance. In Table \ref{tab:rules}, we take the difference of the average of number of rules over all the other classifiers and the proposed classifier in the last column. It is found that the difference is substantial which essentially shows the significance of the proposed pruning strategy. We also compute the percentage decrease of the number of rules on average in Table \ref{tab:rules}. Furthermore, SigD2 is found to outperform RIPPER in terms of accuracy for most of the datasets, however, RIPPER obtains less rules comparatively. This is majorly because RIPPER greedily modifies the generated rules using the Minimum Discription Length (MDL) principle. RIPPER produces a kind of superset of rules covering all information required for classification in the form of intervals. This indicates that there is potential for further improvements.

Furthermore, ACboost is said to be explainable as the base model called wSigDirect produces meaningful and readable rules. The ensemble model helps in determining the attributes which are of most indicative to determine a class. Consider the example of mushroom dataset, the rule produced will be in the format -: (habitat = leaves) and (cap-color = white)  $\rightarrow$ (class = poisonous), where feature name 'habitat' has value 'leaves' and feature name 'cap-color' has value equal to 'white'. This rule along with other similar rules can be further used in the classification phase to determine whether a mushroom is poisonous or not. Similarly for ACbag, the readable rules from the base classifiers can help in interpreting the results. 
\begin{table*}[t]
\centering
\captionsetup[subtable]{position = top}
\caption{Best and runner-up counts comparison from (a) Table 1 and (b) Table 2 on the basis of classification accuracy}
\label{tab:bestrunnerup}
\hspace*{-5em}
\begin{subtable}{0.25\linewidth} 
\centering
\caption{}
\begin{scriptsize}
\begin{tabular}{|c|c|c|c|}
\hline 
\textbf{Classifiers}&\textbf{Best}&\textbf{Runner-up}\\
\hline  
C4.5&2&0\\
\hline
RIPPER&5&2\\
\hline 
CBA&1&1\\
\hline 
CMAR&3&1\\
\hline 
CPAR&1&2\\
\hline 
SigDirect&2&5\\
\hline
SigD2&6&4\\
\hline 
\end{tabular}
\end{scriptsize}
\end{subtable}%
\hspace*{5em}
\begin{subtable}{0.3\linewidth}
\centering
\caption{}
\begin{scriptsize}
\begin{tabular}{|c|c|c|c|}
\hline 
\textbf{Classifiers}&\textbf{Best}&\textbf{Runner-up}\\
\hline 
SVM&0&1\\
\hline  
ANN&4&1\\
\hline
DNN&3&2\\
\hline
SigDirect&1&0\\
\hline 
SigD2&1&2\\
\hline 
ACboost&9&3\\
\hline 
ACbag&1&6\\
\hline
\end{tabular}
\end{scriptsize}
\end{subtable}
\end{table*}
\subsection{Statistical Analysis}
For better understanding the performance over various datasets, we use Demsar's method \cite{demvsar2006statistical} to perform statistical tests in order to compare different algorithms over different datasets. We perform non parametric Friedman's test for comparing the contenders with the proposed approaches. Friedman's test is generally used to compare more than two samples that are related.
The default assumption is that the samples have the same distribution. The assumption is rejected if the probability of observing the data samples given the base assumption(p-value) exceeds the significance threshold value (alpha). The Friedman's test is said to give significant results when the p-value is less than alpha. After analysis of Friedman's test with algorithms in Table 1 and Table 2, we obtained a p-value which is less than alpha (=0.05), which shows that at least one of the samples is significantly different from other samples. Hence, the results are found to be statistically significant. 

Furthermore, we also perform Wilcoxon's signed-ranks test which is another non-parametric statistical hypothesis test to compare the performances of proposed algorithms and the contenders in a pairwise manner. 
%
In this test, initially the differences in the results obtained from the considered pair of algorithms is evaluated for all the datasets to calculate the absolute differences and further to sign each rank. For all the cases where tie in ranks occurs, the average rank is calculated. All the cases with the difference value of zero are ignored.
If the original difference is positive then the rank remains positive, however, if the difference is found to be less than zero, then the rank is multiplied by -1. Further, the sum of positive and negative ranks are used to calculate the z-score values as defined in \cite{demvsar2006statistical}. With the z-score value smaller than -1.96, the corresponding p-value is less than 0.05 which leads to the rejection of the null hypothesis.

The results in Table \ref{tab:pvalue} show that, SigD2 is significantly better than C4.5, CBA, CMAR, CPAR and SVM. However, the performance when compared with the original SigDirect seems to be quite similar and the p-value comes out to be greater than 0.05. We assume that, although there might not be difference in terms of classification accuracy, however, the new pruning strategy of SigD2 is more substantial and promising as it has reduced the number of rules to a small number as compared to the original SigDirect. SigD2's performance is found to be as good as RIPPER, however, with more wins and a p-value of 0.07. Furthermore, SigD2, ANN and DNN are at par with 7 wins and 7 loses based on classification accuracy.

The results from ACboost are found to be statistically significant than those of SigD2, ANN ,DNN and SVM as p-value is less than the significance level of 0.05. Moreover, ACbag although performs better than SVM and SigD2, it has a comparable performance if not better than ANN, as the p-value is greater than 0.05. Thus, the results obtained in this section highlight the significance of the explainable models over the ones that are hard to interpret (ANN, DNN \& SVM). SigD2 and ACboost are almost at par with other strong learners like neural network in terms of classification accuracy along with its ability to be interpreted using a limited number of rules.

\begin{table}[t]
    \centering
    \caption{Statistical analysis of Table\ref{tab:accuracy_phase1} and Table \ref{tab:accuracy_phase2}}   
    \label{tab:pvalue}
\begin{scriptsize}
\begin{tabular}{|c|c|c|c|c|c|}
\hline 
\textbf{Classifiers}&\textbf{Wins}&\textbf{Losses}&\textbf{Ties}&\textbf{p-value}\\
\hline  
SigD2 vs C4.5*&12&3&0&0.005\\
\hline
SigD2 vs RIPPER&10&4&1&0.074\\
\hline 
SigD2 vs CBA*&13&2&0&0.005\\
\hline 
SigD2 vs CMAR*&11&2&2&0.033\\
\hline 
SigD2 vs CPAR*&12&3&0&0.008\\
\hline 
SigD2 vs SigDirect&10&4&1&0.272\\
\hline 
SigD2 vs SVM*&12&3&0&0.041\\
\hline
SigD2 vs ANN&7&7&1&0.510\\
\hline
SigD2 vs DNN&7&7&1&0.510\\

\Xhline{4\arrayrulewidth}
ACbag vs SigD2*&11&3&1&0.064\\

\hline
ACbag vs SVM*&12&3&0&0.005\\
\hline 
ACbag vs ANN&9&5&1&0.140\\
\hline 
ACbag vs DNN&8&6&1&0.140\\
\hline 
ACboost vs SigD2*&12&2&1&0.002\\
\hline 
ACboost vs SVM*&14&1&0&0.002\\
\hline 
ACboost vs ANN*&10&4&1&0.016\\
\hline 
ACboost vs DNN*&10&4&1&0.022\\
\hline 
\end{tabular}
\begin{tablenotes}
      \small
      \centering
      \item (*) indicates statistically significant results with a p-value of 0.05.
\end{tablenotes}
\end{scriptsize}
\end{table}

\section{Conclusion and Future Work}\label{sec:conclusion}
In this paper, we present a competitive associative classifier, which builds a rule-based model that is explainable, readable and minimalist. The classifier initially performs a rule generation step followed by a two phase rule pruning step to obtain the classification rules. The proposed rule pruning strategy reduces the rule set to a significantly small number, making it more useful for various applications especially in the field of bio-medicine where model interpretability is important. The proposed approaches are at par with the other supervised classifiers like ANN and SVM, which are black boxes and do not provide interpretable classification models. Unlike them, SigD2 is an explainable classifier. 
Furthermore, ACboost algorithm uses an ensemble of wSigDirect, to build a strong learner that boosts the prediction performance.
%

The results obtained are very encouraging; we would like to work on making the classifier more efficient in terms of rule generation phase. Although, we are able to compete with RIPPER in terms of accuracy, the number of rules for RIPPER are still smaller. In future, we intend to identify rules that are noisy and can be potentially removed. We also intend to use our proposed approach on various health-care related applications where explanation of prediction is required. Furthermore, since SigD2 produces human readable rules, we would like to study the possibility of injecting human expert knowledge to the obtained rules in order to further improve the prediction performance.
  
\section{Acknowledgment}\label{sec:ackwdgmnt}
We would like to acknowledge the contributions of Mohammad-Hossein Motallebi for implementing the original SigDirect and Parnian Yousefi for discussions on Ripper.

\bibliographystyle{splncs04}
\bibliography{ref}
\end{document}